\documentclass{article}
\usepackage[utf8]{inputenc}

\usepackage{multicol}
\usepackage{multirow}
\usepackage{longtable}
\usepackage{array}
\usepackage{rotating}
\usepackage{amsmath}

\usepackage{lastpage,fancyhdr,graphicx}

\title{Classification in biological networks with hypergraphlet kernels}
\author{Jose Lugo-Martinez\, and Predrag Radivojac \\
Department of Computer Science \\ Indiana University \\ 
Bloomington, Indiana 47405, U.S.A.}

\date{}

\begin{document}

\maketitle

\begin{abstract}
Biological and cellular systems are often modeled as graphs in which vertices represent objects of interest (genes, proteins, drugs) and edges represent relational ties among these objects (binds-to, interacts-with, regulates). This approach has been highly successful owing to the theory, methodology and software that support analysis and learning on graphs. Graphs, however, often suffer from information loss when modeling physical systems due to their inability to accurately represent multiobject relationships. Hypergraphs, a generalization of graphs, provide a framework to mitigate information loss and unify disparate graph-based methodologies. In this paper, we present a hypergraph-based approach for modeling physical systems and formulate vertex classification, edge classification and link prediction problems on (hyper)graphs as instances of vertex classification on (extended, dual) hypergraphs in a semi-supervised setting. We introduce a novel kernel method on vertex- and edge-labeled (colored) hypergraphs for analysis and learning. The method is based on exact and inexact (via hypergraph edit distances) enumeration of small simple hypergraphs, referred to as hypergraphlets, rooted at a vertex of interest. We extensively evaluate this method and show its potential use in a positive-unlabeled setting to estimate the number of missing and false positive links in protein-protein interaction networks. \\
\end{abstract}

\section{Introduction}

Graphs provide a mathematical structure for describing relationships between objects in a system. Owing to their intuitive representation, well-understood theoretical properties, the wealth of the algorithmic methodology and available code base, graphs have also become a major framework for modeling biological systems. Protein-protein interaction networks, protein 3D structures, drug-target interaction networks, metabolic networks and gene regulatory networks are some of the major representations of biological systems. Unfortunately, molecular and cellular systems are only partially observable and may contain significant amount of noise due to their inherent stochastic nature as well as the limitations of both low-throughput and high-throughput experimental techniques. This highlights the need for the development and application of computational approaches for predictive modeling (e.g., inferring novel interactions) and identifying interesting patterns in such data.

Learning on graphs can be generally seen as supervised or unsupervised. Under a supervised setting, typical tasks involve \emph{graph classification}; i.e., the assignment of class labels to entire graphs~\cite{R-T2010}, \emph{vertex or edge classification}; i.e., the assignment class labels to vertices or edges in a single graph~\cite{R-K2009}, or \emph{link prediction}; i.e., the prediction of the existence of edges in graphs~\cite{R-Li2007}. Alternatively, frequent subgraph mining ~\cite{R-J2013}, motif finding~\cite{Milo2002}, clustering~\cite{Agarwal2006}, and community detection~\cite{R-F2010} are traditional unsupervised approaches. Regardless of the category, the development of techniques that capture local/global network structure, measure graph similarity and incorporate domain-specific knowledge in a principled manner lie at the core of all these problems. 

The focus of this study is on classification problems across various biological networks. A straightforward approach to this problem is the use of topological and other descriptors (e.g., vertex degree, clustering coefficient, betweenness centrality) that summarize graph neighborhoods. These descriptors straightforwardly lead to vector-space representations of vertices or edges in the graph, after which standard machine learning algorithms can be applied to learn a target function~\cite{Chung1997, R-X2006}. Another approach involves the use of kernel functions on graphs~\cite{Vishwanathan2010}. Kernels are mappings of pairs of objects from an input space $\mathcal{X}$ to an output space $\mathcal{Y}$ with special properties, such as symmetry and positive semi-definiteness, that lead to efficient learning. Graph kernels often exploit similar ideas as traditional vector-space approaches. Finally, classification on graphs can be approached using probabilistic graphical models such as Markov Random Fields~\cite{R-K2009} and related label-propagation~\cite{R-Z2002} or flow-based~\cite{R-N2005} methods. These ``global'' formulations are generally well adjusted to learning smooth functions over neighboring nodes.

Despite the success and wide adoption of these methods in machine learning and computational biology, it is well-understood that graph representations suffer from information loss since every edge can only encode pairwise relationships~\cite{Klamt2009}. A protein complex, for instance, cannot be distinguished from a set of proteins that interact only pairwise. Such disambiguation, however, is important in order to understand the biological activity of these molecules. Hypergraphs, a generalization of graphs, naturally capture these higher-order relationships~\cite{Berge1973}. As we show later, they also provide a representation that can be used to unify several conventional classification problems on (hyper)graphs as a single vertex classification approach on hypergraphs.

In this paper, we present and evaluate a kernel-based framework for the problems of vertex classification, edge classification and link prediction in graphs and hypergraphs. We first use the concepts of hypergraph duality to demonstrate that all such classification problems can be unified through the use of hypergraphs. We then describe the development of edit-distance hypergraphlet kernels for vertex classification in hypergraphs and combine them with support vector machines into a semi-supervised predictive methodology. Finally, we use sixteen biological network data sets, eleven assembled specifically for this work, to provide evidence that the proposed approaches compare favorably to the previously established methods.

\section{Background}

\subsection{Graphs and hypergraphs}
\label{sec:graphs}

\textbf{Graphs.} A graph $G$ is a pair $(V, E)$, where $V$ is a set of vertices (nodes) and $E \subseteq V \times V$ is a set of edges. In a vertex-labeled graph, a labeling function $f$ is defined as $f : V \rightarrow \Sigma$, where $\Sigma$ is a finite alphabet. Similarly, in an edge-labeled graph, another labeling function $g$ is given as $g : E \rightarrow \Xi$, where $\Xi$ is also a finite set. %We refer to $\Sigma$ and $\Xi$ as the vertex-label and edge-label alphabets, respectively. 
A rooted graph $G$ is a graph together with one distinguished vertex called the root. We denote such graphs as $G = (V, v, E)$, where $v \in V$ is the root. A neighborhood graph $N_{n-1}(v)=(V(v), v, E(v))$ of a vertex $v \in V$ is a rooted graph constructed from $G$ such that all nodes at distance at least $n$ from $v$ (and corresponding edges) are removed.

In this work we focus on undirected (the order of the vertices in each edge can be ignored), simple graphs (graphs without self-loops). Additionally, for the simplicity of presentation, we ignore weighted graphs; i.e., graphs where a non-negative number is associated with each vertex. Generalization of our approach and terminology to directed and weighted graphs is straightforward.

A walk $w$ of length $k$ in a graph $G$ is a sequence of nodes $v_1, \cdots, v_k$ such that $(v_{i}, v_{i+1}) \in E$, for $1 \leq i < k$. If $v_1=v_k$, $w$ is called a cycle of length $k-1$. A path $p$ in $G$ is a walk in which all nodes are distinct. A connected graph is a graph where there is a path between any two nodes.

\textbf{Hypergraphs.} A hypergraph $G$ is a pair $(V, E)$, where $V$ is the vertex set as previously defined and $E$ is a family of non-empty subsets of $V$ called hyperedges. As in the case of graphs, one can define a vertex-labeled, edge-labeled, rooted, and neighborhood hypergraphs. A hyperedge $e$ is said to be incident with a vertex $v$ if $v \in e$ and two vertices are called adjacent if there is an edge that contains both vertices. The neighbors of a vertex $v$ in a hypergraph are the vertices adjacent to $v$. Two hyperedges are said to be adjacent if their intersection is non-empty. Finally, the degree $d(v)$ of a vertex $v$ in a hypergraph is given by $d(v) = |\{e \in E  \, | \, v \in e\}|$, whereas the degree $\delta(e)$ of a hyperedge $e$ is defined as its cardinality; that is, $\delta(e) = |e|$. 

A walk $w$ of length $k$ in a hypergraph $G=(V,E)$ is a sequence of vertices and hyperedges $v_1, e_1, \cdots, e_{k-1}, v_{k}$ such that $(v_{i}, v_{i+1}) \in e_i$ for each $1 \leq i < k$ and $e_i \in E$. If $v_1=v_k$, $w$ is called a cycle of length $k-1$. A path $p$ in a hypergraph is a walk in which all nodes and edges are distinct. A connected hypergraph is a hypergraph where there exists a path between any two nodes.

\textbf{Isomorphism.} Consider two graphs, $G = (V, E)$ and $H=(W,F)$. We say that $G$ and $H$ are isomorphic, denoted as $G \cong H$, if there exists a bijection $f : V \rightarrow W$ such that $(u,v) \in E$ if and only if $(f(u), f(v)) \in F$ for all $u,v \in V$. If $G$ and $H$ are hypergraphs, an isomorphism is defined as interrelated bijections $f : V \rightarrow W$ and $g : E \rightarrow F$ such that $e = \{v_1, \cdots, v_{\delta(e)}\} \in E$ if and only if $g(e) = \{f(v_1), \cdots, f(v_{\delta(e)})\} \in F$ for all hyperedges $e \in E$. Isomorphic graphs (hypergraphs) are structurally identical. An automorphism is an isomorphism of a graph (hypergraph) to itself.

\textbf{Edit distance.} Consider two vertex- and hyperedge-labeled hypergraphs $G$ and $H$. The edit distance between these hypergraphs corresponds to the minimum number of edit operations necessary to transform $G$ into $H$, where edit operations are defined as insertion/deletion of vertices/hyperedges and substitutions of vertex and hyperedge labels. Any sequence of edit operations that transforms $G$ into $H$ is referred to as an edit path; hence, the hypergraph edit distance between $G$ and $H$ corresponds to the length of the shortest edit path between them. This concept can be generalized to the case where each edit operation is assigned a cost. Hypergraph edit distance then corresponds to the edit path of minimum cost. 

%Figure 1
\begin{figure}[!tp]
\includegraphics[width=0.9\columnwidth]{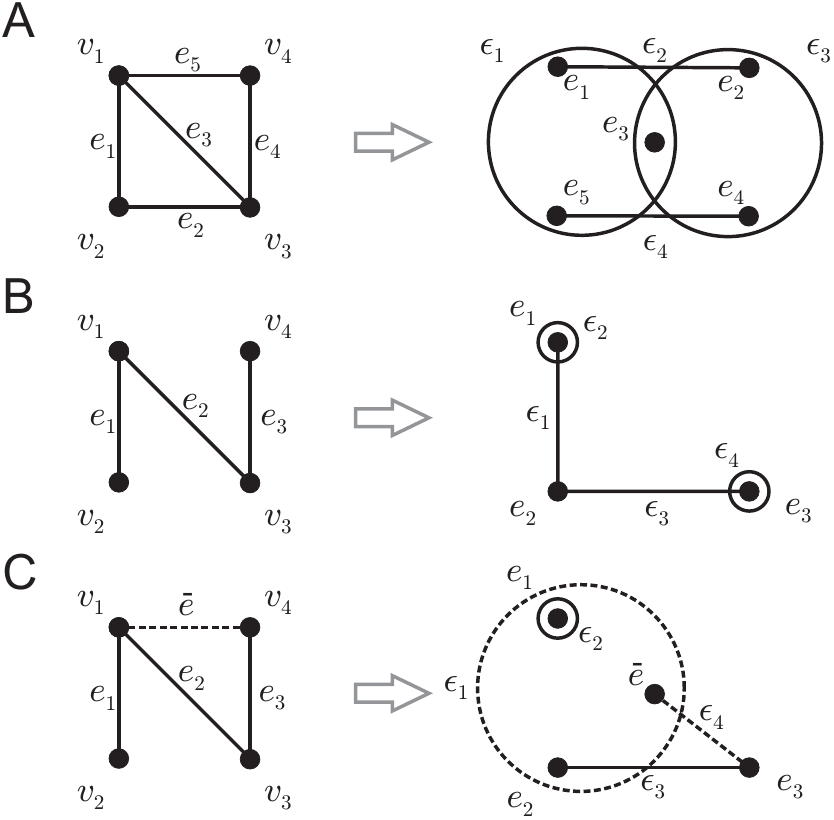}
\caption{\textbf{Examples of hypergraph duality.} Panel (A) shows a hypergraph $G=(V,E)$, where $V=\{v_1,v_2,v_3,v_4\}$ and $E=\{e_1,$ $e_2,e_3,e_4,e_5\}$ with its dual hypergraph $G^{*}=(V^{*},E^{*})$, where $V^{*}=\{e_1,e_2,e_3,e_4,e_5\}$ and $E^{*}=\{\epsilon_1,\epsilon_2,\epsilon_3,\epsilon_4\}$ such that $\epsilon_1 = \{e_1,e_3,e_5\}$, $\epsilon_2 = \{e_1,e_2\}$, $\epsilon_3 = \{e_2,e_3,e_4\}$ and $\epsilon_4 = \{e_4,e_5\}$. Panel (B) shows an example of graph $G$ with two degree-one vertices that lead to the dual hypergraph $G^*$ with self-loops; $\epsilon_2$ and $\epsilon_4$. Panel (C) shows an extended dual hypergraph that is proposed to formulate link prediction as an instance of vertex classification in hypergraphs. To make a prediction regarding the existence of edge $\bar{e}$, shown as a dashed line on the left side, an extended dual hypergraph is created in which $\bar{e}$ is added to the set of vertices $V^*$. Updates are made to hyperedges $\epsilon_1$ and $\epsilon_4$ (dashed) that correspond to those vertices in $G$ that are incident with the edge $\bar{e}$.}
\label{duality:example}
\end{figure}

\subsection{Hypergraph duality}
\label{duality}
%The concept of hypergraph duality is defined as follows. 
Let $G=(V,E)$ be a hypergraph, where $V=\{v_1,\ldots,v_n\}$ and $E=\{e_1,\ldots,e_m\}$. The dual hypergraph of $G$, denoted as $G^{*}=(V^{*},E^{*})$, is obtained by constructing the set of vertices as $V^{*}=\{e_1,\ldots,e_m\}$ and the set of hyperedges as $E^{*}=\{\epsilon_1,\ldots,\epsilon_n\}$ such that $\epsilon_i = \{e_j\,|\, v_i \in e_j\}$. Figure \ref{duality:example}A-B shows two examples of a hypergraph $G$ and its dual hypergraph representation $G^{*}$. Observe that the hyperedges of the original hypergraph $G$ are the vertices of the dual hypergraph $G^{*}$, whereas the hyperedges of $G^{*}$ are constructed using the hyperedges of $G$ that are incident with the respective vertices.

\subsection{Classification on hypergraphs}
We are interested in binary classification on hypergraphs. The following paragraphs briefly define three distinct classification problems, formulated here so as to naturally lead to the methodology proposed in the next section.

\textbf{Vertex classification.} Given a set of rooted hypergraphs $\mathcal{H}=\left\{ H_{i}\right\} _{i=1}^{n}$, where each $H_{i}=(V,v_{i},E)$ corresponds to the same, possibly disconnected, hypergraph $G=(V,E)$ rooted at a different vertex of interest $v_{i} \in V$. Here, one aims to learn some classifier function $t:\mathcal{H}\rightarrow\left\{ -1,+1\right\}$ using a labeled training set $\mathcal{T} =\left\{ \left(H_{j},t_{j}\right)\right\} _{j=1}^{m}$, where $m < n$, as a means of assigning class labels to each unlabeled vertex in $\mathcal{H}$. A number of classical problems in computational biology map straightforwardly to vertex classification; e.g., protein function prediction~\cite{Sharan2007}, disease gene prioritization~\cite{Moreau2012}, and so on.

\textbf{Hyperedge classification.} Given a possibly disconnected hypergraph $G=(V,E)$, the objective is to learn a discriminant function $t: E \rightarrow\left\{-1,+1\right\}$ from a labeled training set $\mathcal{T} =\left\{ \left(e_{i},t_{i}\right)\right\} _{i=1}^{m}$, where $m < |E|$, and infer class annotations for every unlabeled hyperedge in $E$. An example of edge classification is the prediction of types of macromolecular interactions such as positive vs. negative regulation.

\textbf{Link prediction.} Let $G=(V,E)$ be a hypergraph with some missing hyperedges and let $\bar{E}$ be all non-existent hyperedges in $G$; i.e., $\bar{E} = \mathcal{U} - E$, where $\mathcal{U}$ represents all possible hyperedges over $V$. The goal is to learn a target function $t: \mathcal{U}\rightarrow\left\{-1,+1\right\}$ and infer the existence of all missing hyperedges. Examples of link prediction include predicting protein-protein interactions, predicting drug-target interactions, and so on.

\subsection{Positive-unlabeled learning}
\label{positive:unlabeled:learning}
A number of prediction problems in computational biology can be considered within a semi-supervised framework, where a set of labeled and a set of unlabeled examples are used to construct classifiers that discriminate between positive and negative examples. A special category of semi-supervised learning occurs when labeled data contain only positive examples; i.e., where the negative examples are either unavailable or ignored; say, if the set of available negatives is small or biased. Such problems are generally referred to as learning from positive and unlabeled data or positive-unlabeled learning~\cite{Denis2005}. Many prediction problems in molecular biology belong to the open world category; i.e., due to various experimental reasons, the absence of evidence of class labels is not the evidence of absence. Such problems lend themselves naturally to the positive-unlabeled setting.

Research in machine learning has recently established tight connections between traditional supervised learning and (non-traditional) positive-unlabeled learning. Under mild conditions, a classifier that optimizes the ranking performance; e.g., area under the ROC curve~\cite{Fawcett2006}, in the non-traditional setting has been shown to also optimize the performance in the traditional setting~\cite{Elkan2008, Blanchard2010, Menon2015}. Similar relationships have been established in approximating posterior distributions~\cite{Jain2016b, Jain2016} as well as in recovering the true performance accuracy in the traditional setting for a classifier evaluated in a non-traditional setting~\cite{Jain2017}. The latter two problems require estimation of class priors; i.e., the fractions of positive and negative examples in (representative) unlabeled data~\cite{Jain2016b, Jain2016, Ramaswamy2016}.

\section{Methods}

\subsection{Problem formulation}
\label{sec:problem}
We consider binary classification problems on graphs and hypergraphs and propose to unify all such learning problems through semi-supervised vertex classification on hypergraphs. First, vertex classification falls trivially into this framework. Second, the problems of edge classification in graphs and hyperedge classification in hypergraphs are equivalent to the problem of vertex classification on dual hypergraphs. As discussed in Section~\ref{duality}, both graphs and hypergraphs give rise to dual hypergraph representations and, thus, (hyper)edge classification on a graph $G$ straightforwardly translates into vertex classification on its dual hypergraph $G^*$. We note here that vertices with the degree of one in $G$ give rise to self-loops in the dual hypergraph $G^*$. To account for them, we add one dummy node per self-loop with the same vertex label as the original vertex and connect them with an appropriately labeled edge. Third, one can similarly see link prediction as vertex classification on dual hypergraphs, where the set of existing links is treated as positive data, the set of known non-existing links is treated as negative data, and the remaining set of missing links is treated as unlabeled data. This formulation further requires an extension of dual hypergraph representations as follows. Consider a particular negative or missing link $\bar{e} \in \bar{E}$ in the original graph $G$ with its dual hypergraph $G^*$ (Fig.~\ref{duality:example}C). To make a prediction on this edge $\bar{e}$, we must first introduce a new vertex $\bar{e}$ in the dual hypergraph as well as modify those hyperedges in $G^*$ that correspond to the vertices $v \in \bar{e}$ in $G$ (Fig.~\ref{duality:example}C). We denote this extended hypergraph as $G^{*}_{\bar{e}}$. It now easily follows that the sets of negative and unlabeled examples can be created by considering a collection of extended graphs $G^{*}_{\bar{e}}$, one at a time, for all non-existing vertices $\bar{e} \in \bar{E}$ or a subset thereof. 

Since most graph data in biological networks lack large sets of representative negative examples, we approach vertex classification, (hyper)edge classification and link prediction as instances of vertex classification on (extended, dual) hypergraphs in a positive-unlabeled setting.  We believe this is a novel and useful attempt at generalizing three distinct graph classification problems in a common kernel-based semi-supervised setting. The following sections introduce hypergraphlet kernels that are the core of our classification approach.

%Figure 2
\begin{figure}[!tp]
\includegraphics[width=.85\columnwidth]{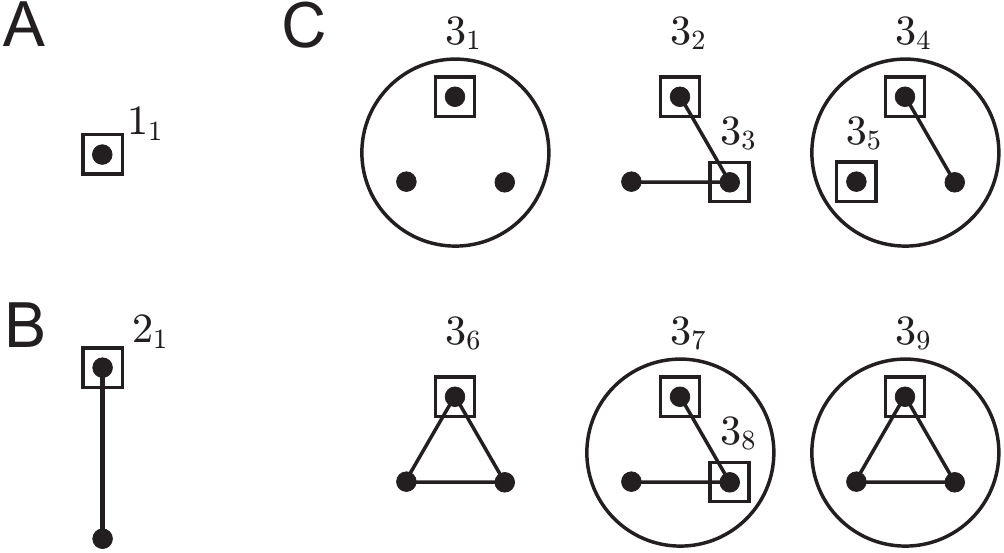}
\caption{\textbf{Undirected base hypergraphlets.} All undirected hypergraphlets with $1$ to $3$ vertices, with the root node of each hypergraphlet inscribed in a square. All hypergraphlets are presented in a compressed notation; e.g., the two non-isomorphic hypergraphlets $3_2$ and $3_3$ are shown in one drawing.} 
\label{hypergraphlets}
\end{figure}

\subsection{Hypergraphlets}

\textbf{Hypergraphlets.} Inspired by graphlets~\cite{Przulj2004, Przulj2007}, we define \emph{hypergraphlets} as small, simple, connected, rooted hypergraphs. A hypergraphlet with $n$ vertices is called an $n$-hypergraphlet; and the $i$-th hypergraphlet of order $n$ is denoted as $n_i$. We consider hypergraphlets up to isomorphism and will refer to these isomorphisms as root- and label-preserving isomorphisms when hypergraphs are rooted and labeled. Figure~\ref{hypergraphlets} displays all non-isomorphic unlabeled $n$-hypergraphlets with up to three vertices. There is only one hypergraphlet of order $1$ ($1_1$; Fig.~\ref{hypergraphlets}A) and one hypergraphlet of order $2$ ($2_1$; Fig.~\ref{hypergraphlets}B). On the other hand, there are nine hypergraphlets of order $3$ ($3_1,\ldots,3_9$; Fig.~\ref{hypergraphlets}C) and $461$ hypergraphlets of order $4$ (not shown). We refer to all these hypergraphlets as \emph{base hypergraphlets} since they correspond to the case when $|\Sigma|=|\Xi|=1$. 

Consider now a vertex- and hyperedge-labeled (or \emph{fully labeled} for short) hypergraphlet with $n$ vertices and $m$ hyperedges, where $\Sigma$ and $\Xi$ denote the vertex-label and hyperedge-label alphabets, respectively. If $|\Sigma| > 1$ and/or $|\Xi| > 1$, then automorphic structures with respect to the same base hypergraphlet may exist; hence, the number of fully labeled hypergraphlets per base structure is generally smaller than $|\Sigma|^{n} \cdot |\Xi|^{m}$. For example, if one only considers vertex-labeled $3$-hypergraphlets, then there are $|\Sigma|^3$ vertex-labeled hypergraphlets corresponding to the asymmetric base hypergraphlets $3_2$, $3_4$ and $3_7$ but only $\frac{1}{2} (|\Sigma|^3 + |\Sigma|^2)$ corresponding to the base hypergraphlets $3_1$, $3_3$, $3_5$, $3_6$, $3_8$, $3_9$; see Table~\ref{hypergraphs:polya}. 
This is a result of symmetries in the base hypergraphlets that give rise to automorphisms among vertex-labeled structures. Similarly, if $|\Xi| > 1$, then new symmetries may exist with respect to the base hypergraphlets that give rise to different automorphisms among hyperedge-labeled structures. In Section~\ref{growth:pattern}, we provide a more detailed discussion on these symmetries. 
The relevance of these symmetries and enumeration steps relates to the dimensionality of the Hilbert space in which the prediction is carried out. 

\subsection{Hypergraphlet kernels}
Motivated by the case for graphs~\cite{Shervashidze2009,Vacic2010,Lugo2014}, we introduce \emph{hypergraphlet kernels}. Let $G=(V, E, f, g, \Sigma, \Xi)$, be a fully labeled hypergraph where $f$ is a vertex-labeling function $f : V \rightarrow \Sigma$, $g$ is a hyperedge-labeling function $g : E \rightarrow \Xi$, and $|\Sigma|, |\Xi| \geq 1$. The vertex- and hyperedge-labeled $n$-hypergraphlet count vector for any vertex $v \in V$ is defined as 

\begin{equation}
\phi_{n} (v) = (\varphi_{n_1} (v), \varphi_{n_2}(v), \ldots ,\varphi_{n_{\kappa(n, \Sigma, \Xi)}} (v)),
\end{equation}

\noindent where $\varphi_{n_i} (v)$ is the count of the $i$-th fully labeled $n$-hypergraphlet and $\kappa(n, \Sigma, \Xi)$ is the total number of vertex- and hyperedge-labeled $n$-hypergraphlets. A kernel function between the $n$-hypergraphlet counts for vertices $u$ and $v$ is defined as an inner product between $\phi_n (u)$ and $ \phi_n (v)$; i.e.,

\begin{equation}
k_n (u, v)= \left\langle \phi_n (u), \phi_n (v) \right\rangle .
\label{eq:kernel}
\end{equation}

\noindent The hypergraphlet kernel function incorporating all hypergraphlets up to the size $N$ is given by 

\begin{equation}
k(u, v) = \sum_{n = 1}^{N} k_n (u, v),
\label{eq:combined:kernel}
\end{equation}

\noindent where $N$ is a small integer. In this work we use $N=4$ due to the exponential growth of the number of base hypergraphlets. 

\subsection{Edit-distance hypergraphlet kernels}
Consider a fully labeled hypergraph $G=(V, E, f, g, \Sigma, \Xi)$. Given a vertex $v \in V$, we define the vector of counts for a $\tau$-generalized edit-distance hypergraphlet representation as

\begin{equation}
\phi_{(n, \tau)} (v) = (\psi_{(n_1, \tau)} (v), \psi_{(n_2, \tau)} (v), \ldots, \psi_{(n_{\kappa(n, \Sigma, \Xi)}, \tau)} (v)),
\end{equation}

\noindent where

\begin{equation}
\psi_{(n,\tau)}(v)=\sum_{n_{j}\in E(n_{i},\tau)}c(n_{i},n_{j})\cdot\varphi_{n_{j}}(v).
\end{equation}

\noindent Here, $E(n_{i},\tau)$ is the set of all $n$-hypergraphlets such that for each $n_{j}\in E(n_{i},\tau)$ there exists an edit path of total cost at most $\tau$ that transforms $n_{i}$ into $n_{j}$ and $c(n_{i},n_{j}) \geq 0$ is a user-defined constant. In words, the counts for each hypergraphlet $n_i$ are updated by also counting all other hypergraphlets $n_j$ that are in the $\tau$ vicinity of $n_i$. The function $c$ can be used to adjust the weights of these pseudocounts. We set $c(n_{i},n_{j}) = 1$ for all $i$ and $j$ and the cost of all edit operations was also set to $1$. This restricts $\tau$ to nonnegative integers.

The length-$\tau$ edit-distance $n$-hypergraphlet kernel $k_{(n, \tau)} (u, v)$ between vertices $u$ and $v$ can be computed as an inner product between the respective count vectors $\phi_{(n, \tau)}(u)$ and $\phi_{(n, \tau)}(v)$; i.e.,
\begin{equation}
k_{(n, \tau)}(u,v)= \left\langle \phi_{(n, \tau)}(u), \phi_{(n, \tau)}(v) \right\rangle.
\label{eq:kernel2}
\end{equation}
Finally, the length-$\tau$ edit-distance hypergraphlet kernel function is given as
\begin{equation}
k_{\tau}(u, v) = \sum_{n = 1}^{N} k_{(n, \tau)}(u, v).
\label{eq:combined:kernel2}
\end{equation}
\noindent The edit operations considered here incorporate substitutions of vertex labels, substitutions of hyperedge labels, and insertions/deletions (indels) of hyperedges. Given these edit operations, we also define three subclasses of  edit-distance hypergraphlet kernels referred to as vertex label-substitution $k^{vl}_{\tau} (u, v)$, hyperedge label-substitution $k^{hl}_{\tau} (u, v)$ and hyperedge-indel kernels $k^{hi}_{\tau} (u, v)$. 

Although the functions from Equations (\ref{eq:kernel}) and (\ref{eq:kernel2}) are defined as inner products, other formulations such as radial basis functions can be similarly considered~\cite{Shawe2009}. We also note that the combined kernels from Equations (\ref{eq:combined:kernel}) and (\ref{eq:combined:kernel2}) can be generalized beyond linear combinations~\cite{Shawe2009}. For the simplicity of this work, however, we only explore equal-weight linear combinations and normalize the functions from Equations (\ref{eq:kernel}) and (\ref{eq:kernel2}) using a cosine transformation.

\subsection{Computational complexity}
The implementation and the analysis of hypergraphlet kernels is an extension of the available solutions for string kernels~\cite{Rieck2008}. Let $N_{n-1}(v)=(V(v), E(v))$ be a neighborhood hypergraph, as defined in Section~\ref{sec:graphs} and suppose it is significantly smaller than the original hypergraph $G$. The hypergraphlet counting algorithm takes $\mathcal{O}(|E(v)|+d^{n}_{\max})$ steps, where $d_{\max}$ is the maximum degree of a vertex. Similarly, the generation of the minimum cost edit path takes $\mathcal{O}(n(|\Sigma|+|\Xi|)+(n^2|\Xi|))$ per single hypergraphlet edit operation. Therefore, for each vertex $v$ an order of $$\mathcal{O}(\min\left\{ |V(v)|^{n},\kappa(n,\Sigma,\Xi)\right\} (n(|\Sigma|+|\Xi|)+(n^2|\Xi|))^{\tau})$$ operations are necessary, where the $|V(v)|^{n}$ term enumerates possible $n$-hypergraphlets in $N_{n-1}(v)$. Note that the possible number of edges in a hypergraph $|E(v)|$ can be significantly larger than the possible number of edges in a standard graph. Hence, in a practical setting, the edit distance hypergraphlet kernels could greatly benefit from effective sampling techniques or exploitation of special types of hypergraphlets. The proposed implementation for computing hypergraphlet kernel functions is computed in time linear in the number of non-zero elements.

\section{Experiment design}

In this section we summarize classification problems, data sets, and evaluation methodology. The hypergraphlet kernels were evaluated on the problems of edge classification and link prediction, both of which require generation of dual hypergraphs followed by the subsequent vertex classification approach. 

\subsection{Data sets}

\textbf{Protein-protein interaction data.} The protein-protein interaction (PPI) data was used for both edge classification and link prediction. In the context of edge classification, we are given a PPI network where each interaction is annotated as either direct physical interaction or a co-membership in a complex. The objective is to predict the type of each interacting protein pair as physical vs. complex (PC). For this task, we used the budding yeast \textit{S. cerevisiae} PPI network assembled by Ben-Hur and Noble~\cite{Ben-Hur2005}. 

Another important task in PPI networks is discovering whether two proteins interact. Despite the existence of high-throughput experimental methods for determining interactions between proteins, the PPI network data of all organisms is incomplete~\cite{vonMering2002}. Furthermore, high-throughput PPI data contains a potentially large fraction of false positive interactions~\cite{vonMering2002, Hart2006, Lewis2012}. Therefore, there is a continued need for computational methods to help guide experiments for identifying novel interactions. Under this scenario, there are two classes of link prediction algorithms: (1) prediction of direct physical interactions~\cite{Gomez2003, Ramani2003, Martin2005, Ben-Hur2005} and (2) prediction of co-membership in a protein complex~\cite{Zhang2004, Qiu2008}. In this paper, we focused on the former task and assembled nine species-specific data sets comprised solely of direct protein-protein interaction data derived from public databases (BIND, BioGRID, DIP, HPRD, and IntAct) as of January 2017. We considered only one protein isoform per gene and used experimental evidence types described by Lewis \emph{et al.}~\cite{Lewis2012}. Specifically, we constructed link prediction tasks for: (1) bacterium \textit{E. coli} (EC), (2) budding yeast \textit{S. cerevisiae} (SC), (3) nematode worm \textit{C. elegans} (CE), (4) thale cress \textit{A. thaliana} (AT), (5) fruit fly \textit{D. melanogaster} (DM), (6) human \textit{H. sapiens} (HS), (7) fission yeast \textit{S. pombe} (SP), (8) brown rat \textit{R. norvegicus} (RN), and (9) house mouse \textit{M. musculus} (MM). 

\textbf{Drug-target interaction data.} Identification of interactions between drugs and target proteins is an area of growing interest in drug design and therapy~\cite{Yamanishi2008, Wang2013}. In a drug-target interaction (DTI) network, nodes correspond to either drugs or proteins and edges indicate that a protein is a known target of the drug. Here we used DTI data for both edge classification and link prediction. In the context of edge labeling, we are given a DTI network where each interaction is annotated as direct (binding) or indirect, as well as assigned modes of action as activating or inhibiting. The objective is to predict the type of each interaction between proteins and drug compounds. For this task, we derived two data sets: (1) indirect vs. direct (ID) binding derived from MATADOR, and (2) activation vs. inhibition (AI) assembled from STITCH. Under link prediction setting, the learning task is to predict drug-target protein interactions. In particular, we focus on four drug-target classes: (1) enzymes (EZ), (2) ion channels (IC), (3) G protein-coupled receptors (GR), and (4) nuclear receptors (NR); originally assembled by Yamanishi \emph{et al.}~\cite{Yamanishi2008}. Table~\ref{datasets} summarizes all data sets used in this work.

\begin{table}
\begin{center}
\caption{\textbf{Summary of binary classification tasks and data sets.} For each learning problem, we show the number of vertices ($V$) and edges ($E$) in the entire hypergraph, as well as the largest connected component ($V^{\text{lcc}}$, $E^{\text{lcc}}$). We also show the number of positive ($n_+$), negative ($n_-$), or unlabeled ($n_u$) data points.}
\label{datasets}
\begin{tabular}{llllll}
\hline 
%\textbf{Type} & \textbf{Dataset} & $|V|$ & $|E|$ & $n_{+}$ & $n_{-}/n_{u}$ 
\textbf{Type} & \textbf{Dataset} &  &  &  & 
\\\hline\hline
\multicolumn{6}{c}{\textit{Edge classification}} \\ 
&  & $|V|$ & $|E|$ & $n_{+}$ & $n_{-}$ \\
\hline
PPI & PC & $4,761$ & $22,988$ & $10,517$ & $12,471$ \\ \hline
\multirow{4}{*}{DTI} 
 & \multirow{2}{*}{ID} & $544$ drugs & \multirow{2}{*}{$10,436$} & \multirow{2}{*}{$4,284$} & \multirow{2}{*}{$6,152$} \\
    &  & $2,261$ targets &  &  & \\
 & \multirow{2}{*}{AI} & $378$ drugs & \multirow{2}{*}{$1,039$} & \multirow{2}{*}{$249$} & \multirow{2}{*}{$790$} \\
    &  & $267$ targets &  &  & \\
\hline \hline
\multicolumn{6}{c}{\textit{Link prediction}} \\ 
&  & $|V|$ & $|E|$ & $|V^{\text{lcc}}|$ & $|E^{\text{lcc}}|^{a}$ \\ \hline
\multirow{9}{*}{PPI} 
 & EC & $393$ & $391$ & $100$ & $153$ \\
 & SC & $4,690$ & $26,165$ & $4,674$ & $26,156$ \\
 & CE & $3,026$ & $5,163$ & $2,779$ & $5,014$ \\
 & AT & $5,391$ & $12,825$ & $5,063$ & $12,631$ \\
 & DM & $7,193$ & $23,159$ & $7,086$ & $23,101$ \\
 & HS & $10,841$ & $45,386$ & $10,729$ & $45,327$ \\
 & SP & $853$ & $1,197$ & $685$ & $1,092$ \\
 & RN & $526$ & $532$ & $301$ & $388$ \\ 
 & MM & $2,065$ & $2,833$ & $1,590$ & $2,522$ \\
 \hline
\multirow{8}{*}{DTI} 
% & \multirow{2}{*}{DB} & $1,007$ drugs & \multirow{2}{*}{$7.8\cdot10^5$} & $8$ & \multirow{2}{*}{$3,152$} & \multirow{2}{*}{$3,152$} \\
%    &  & $773$ targets &  & $8$ &  & \\
 & \multirow{2}{*}{EZ} & $445$ drugs & \multirow{2}{*}{$2,926$} & \multirow{2}{*}{$809$} & \multirow{2}{*}{$2,556$} \\
    &  & $664$ targets &  &  & \\
 & \multirow{2}{*}{IC} & $210$ drugs & \multirow{2}{*}{$1,476$} & \multirow{2}{*}{$409$} & \multirow{2}{*}{$1,473$} \\
    &  & $204$ targets &  &  & \\
 & \multirow{2}{*}{GR} & $223$ drugs & \multirow{2}{*}{$635$} & \multirow{2}{*}{$240$} & \multirow{2}{*}{$570$} \\
    &  & $95$ targets &  &  & \\
 & \multirow{2}{*}{NR} & $54$ drugs & \multirow{2}{*}{$90$} & \multirow{2}{*}{$42$} & \multirow{2}{*}{$50$} \\
    &  & $26$ targets &  &  & \\
\hline
\end{tabular}
\end{center}
\begin{flushleft}
\small{a The size of $n_+$ and $n_u$ is given by $|E^{\text{lcc}}|$.}
\end{flushleft}
\end{table}

\subsection{Integrating domain knowledge via vertex alphabet}

To incorporate domain knowledge into the PPI networks, we exploited the fact that each vertex (protein) in the graph is associated with its amino acid sequence. Two methods were used to develop vertex alphabet. First, we mapped each protein into a vector of $k$-mer ($k=4$) counts and then applied hierarchical clustering on these count vectors. A result of the clustering step assigned one of the $|\Sigma_{\text{SK}}|$ vertex labels for each node. Second, we used protein sequences to predict their molecular and biological function (Gene Ontology terms) using the FANN-GO algorithm~\cite{Clark2011}. Hierarchical clustering was subsequently used on the predicted term scores to group proteins into $|\Sigma_{\text{GO}}|$ broad functional categories. In the case of DTI data, target proteins were annotated in a similar manner. For labeling drug compounds, we used the chemical structure similarity matrix computed from SIMCOMP~\cite{Hattori2003}, transformed it into a dissimilarity matrix and then applied hierarchical clustering to group compounds into $|\Sigma_{\text{SS}}|$ structural categories. 

\subsection{Evaluation methodology}

For each data set, we evaluated all hypergraphlet kernels by comparing them to two in-house implementations of random walk kernels on hypergraphs. The random walk kernels were implemented as follows: given a hypergraph $G$ and two vertices $u$ and $v$, simultaneous random walks $w_u$ and $w_v$ were generated from $u$ and $v$ using random restarts. However, in contrast to random walks on standard graphs, a random walk in a hypergraph is a two-step process such that at each step one must simultaneously (1) pick hyperedges $e_{u}$ and $e_{v}$ incident with current vertices $u$ and $v$ respectively, and (2) pick destination vertices $u' \in e_{u}$ and $v' \in e_{v}$. This process is repeated until a pre-defined number of steps is reached. In the conventional random walk implementation on hypergraphs, a walk was scored as 1 if the entire sequences of vertex and hyperedge labels between $w_u$ and $w_v$ matched; otherwise, a walk was scored as 0. After 10,000 steps, the scores over all walks were summed to produce a kernel value between $u$ and $v$. In order to construct a random walk similar to the hypergraphlet edit distance approach, a cumulative random walk kernel was also implemented. Here, any match between the labels of vertices $u_i$ and $v_i$, or hyperedges $e_{u_{i}}$ and $e_{v_{i}}$ in the $i$-th step of each walk was scored as 1, while a mismatch was scored as 0. Thus, a walk of length $\ell$ could contribute between $0$ and $\ell$ to the total count. In each of the random walks, the probability of restart was selected from a set $\left\{0.1, 0.2,\ldots,0.5\right\}$ and the result with the highest accuracy is reported. On the PPI data sets we also evaluated the performance of pairwise spectrum kernels~\cite{Ben-Hur2005}. The $k$-mer size was varied from $k\in\{3, 4, 5\}$ and the result with the highest accuracy is reported. Finally, in the case of the edit distance kernels, we computed the set of normalized hypergraphlet kernel matrices $\mathcal{K}$ using $k^{vl}_{\tau} (x_i, x_j)$, $k^{hl}_{\tau} (x_i, x_j)$, $k^{hi}_{\tau} (x_i, x_j)$, and $k_{\tau}(x_i, x_j)$ for all pairs $(x_i, x_j)$ obtained from a grid search over $\tau = \{0, 1\}$, $|\Sigma| = \{4, 8, 16\}$ and $N = \{3, 4\}$. The result with the highest accuracy is reported. 

The performance of each method was evaluated through a 10-fold cross-validation. In each iteration, 10\% of nodes in the network are selected for the test set, whereas the remaining 90\% are used for training. Support vector machine (SVM) classifiers were used to construct all predictors and perform comparative evaluation. We used SVM$^{light}$ with the default value for the capacity parameter~\cite{Joachims2002}. Once each predictor was trained, we used Platt's correction to adjust the outputs of the predictor to the 0-1 range~\cite{Platt1999}. Finally, we estimated the area under the ROC curve (AUC), which plots the true positive rate (sensitivity, $sn$) as a function of false positive rate (1 - specificity, $1 - sp$).

\section{Results}

\begin{table}
\begin{center}
\caption{\textbf{Area under the ROC curve estimates for each kernel method on edge classification data set using $10$-fold cross-validation.} The highest performance for each data set is shown in boldface.}
\label{edge:classification:results}
\begin{tabular}{llll}
\hline
\textbf{Dataset/Method} & \textbf{PC} & \textbf{ID} & \textbf{AI} \\ 
\hline\hline
\multicolumn{4}{c}{Without domain information, $|\Sigma| = 1$} \\
Hypergraphlet kernel ($\tau = 0$) & $0.747$ & $0.586$ & $0.583$ \\
Hypergraphlet kernel ($\tau = 1$) & $0.757$ & $0.587$ & $0.605$ \\
\multicolumn{4}{c}{With domain information, $\Sigma = \{\Sigma_{\text{GO}} \bigcup \Sigma_{\text{SS}}\}$} \\
Random walk & $0.741$ & $0.589$ & $0.808$ \\ 
Cumulative random walk & $0.760$ & \textbf{0.834} & $0.826$ \\
Hypergraphlet kernel ($\tau = 0$) & $0.774$ & $0.715$ & $0.816$ \\
Hypergraphlet kernel ($\tau = 1$) & \textbf{0.781} & $0.736$ & \textbf{0.845} \\ 
\hline
\end{tabular}
\end{center}
\end{table}
\begin{figure*}
\includegraphics[width=0.32\textwidth]{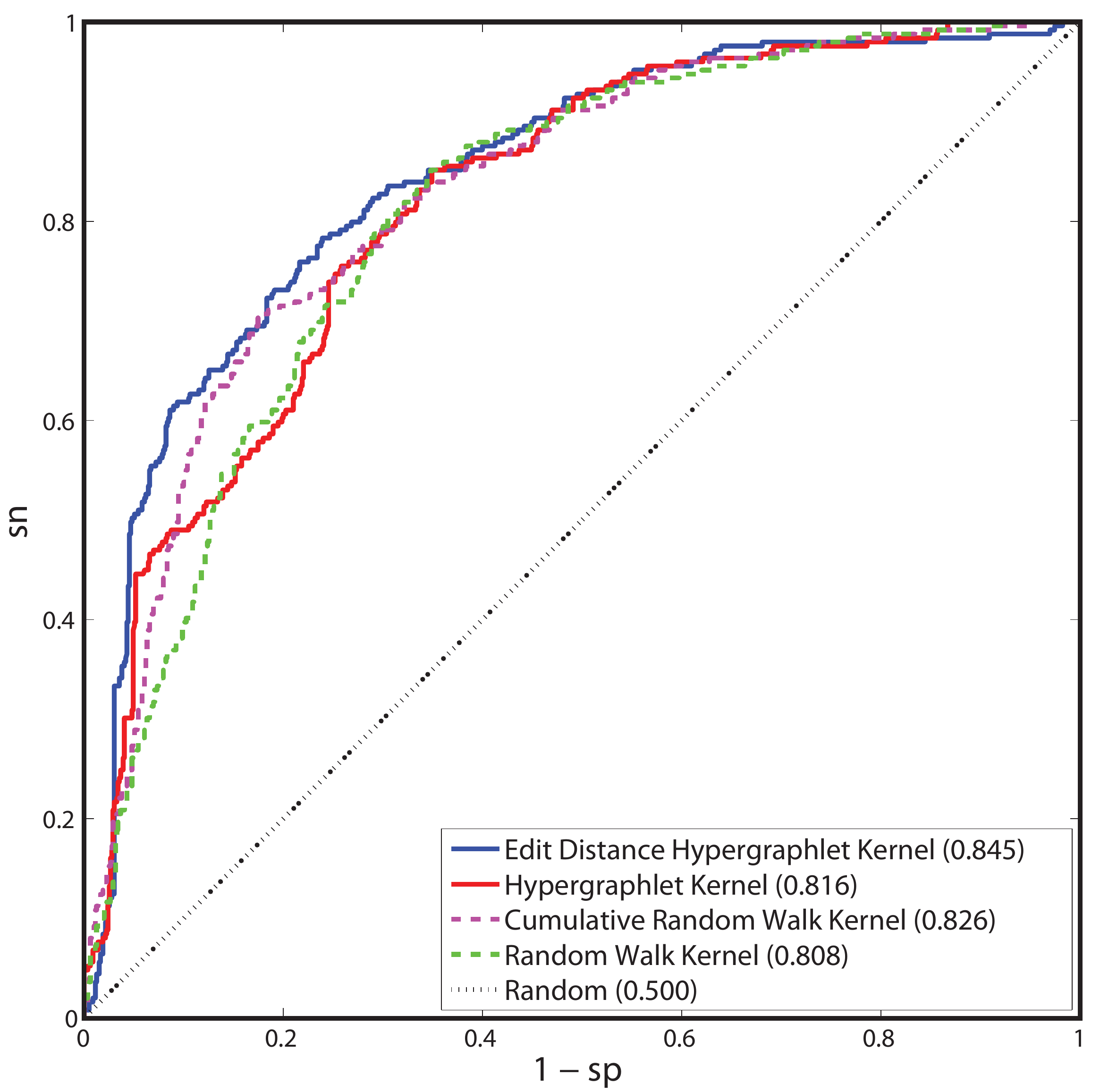} 
\includegraphics[width=0.32\textwidth]{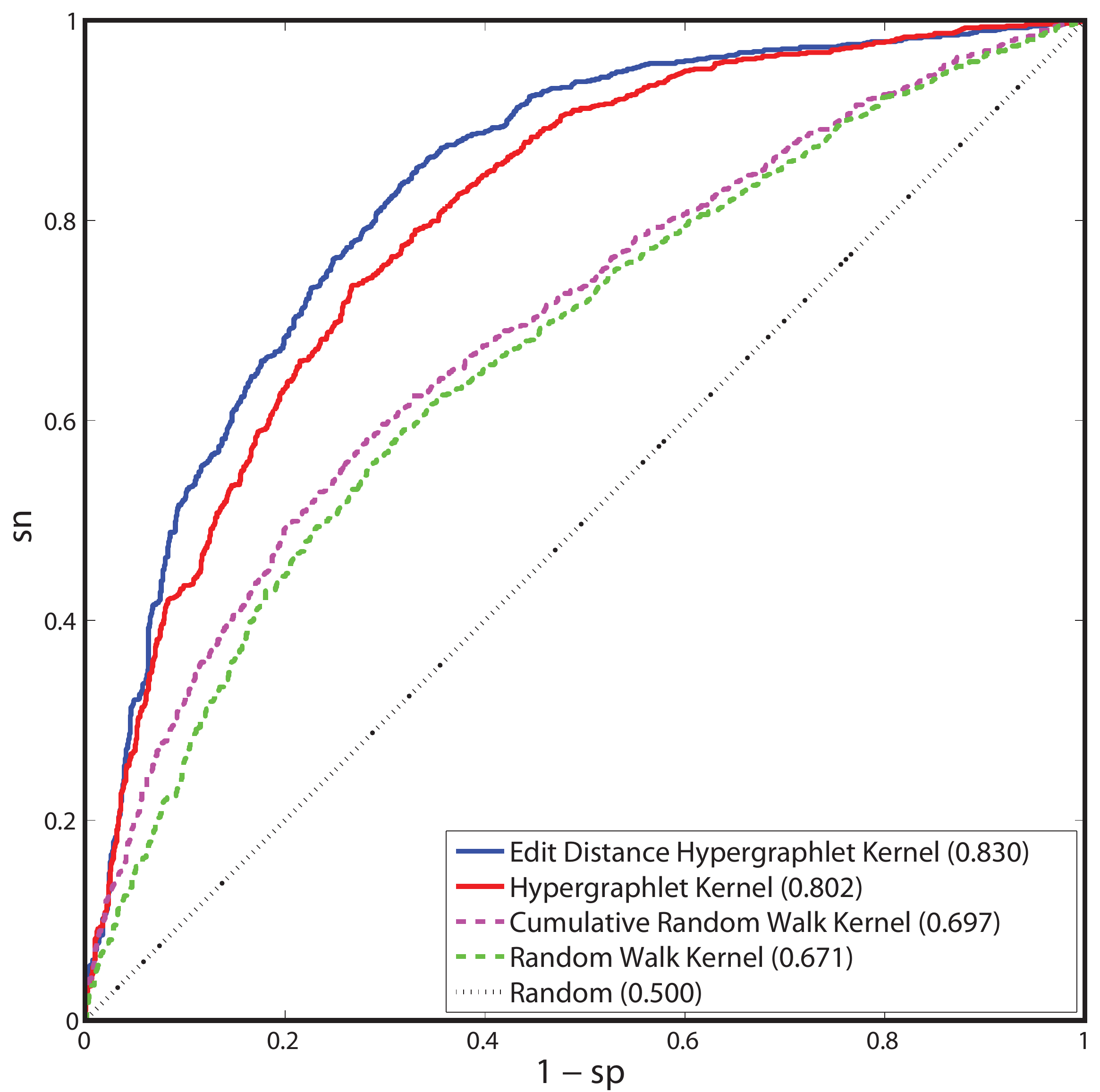} 
\includegraphics[width=0.32\textwidth]{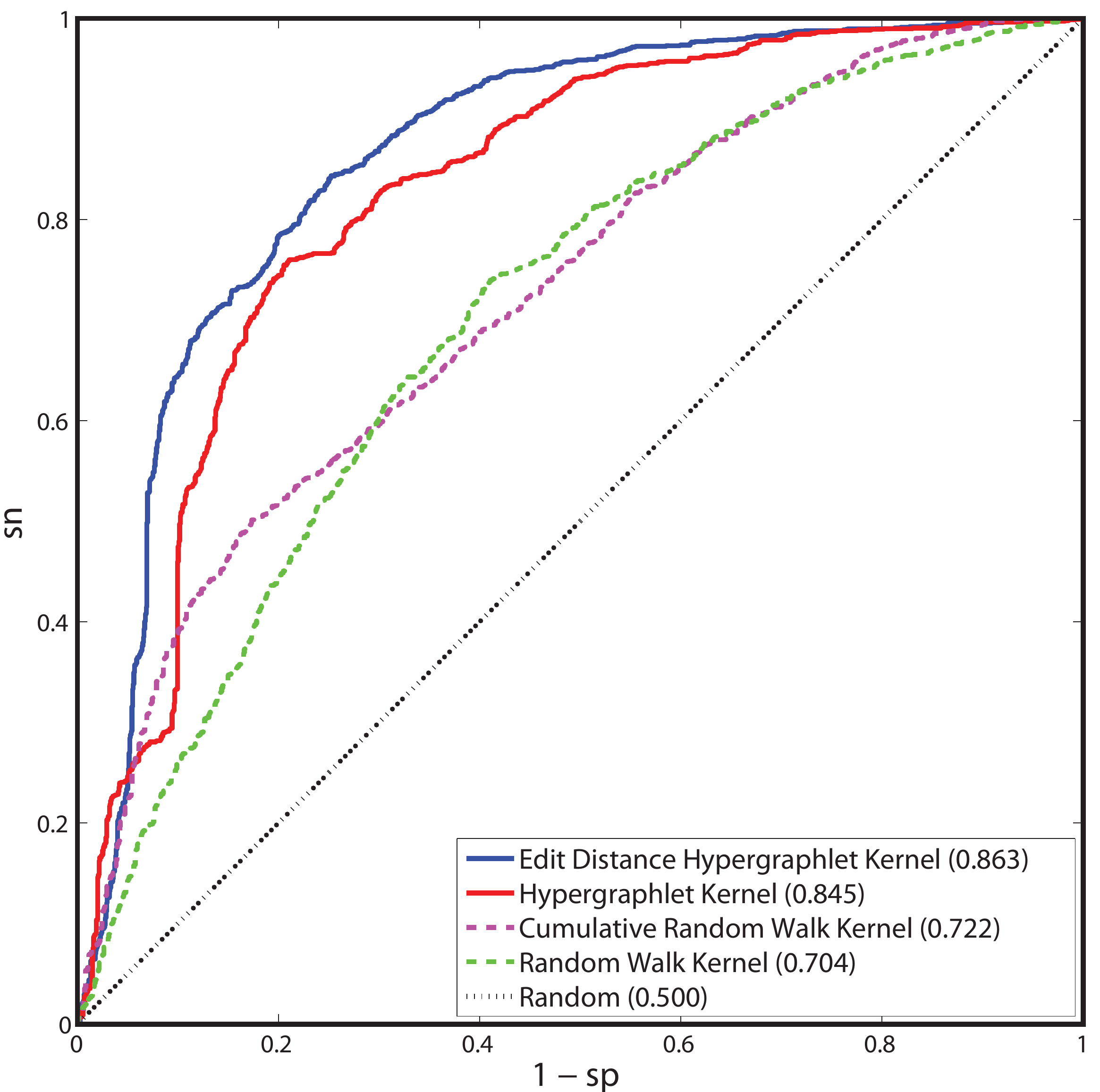} 
\caption{Comparisons of ROC curves between different kernel methods for three representative data sets. (Left) \text{AI}, (Center) \text{SP}, and (Right) \text{IC}. AUC values are shown in parentheses for each method.}
\label{auc:curves}
\end{figure*}

\subsection{Performance analysis on edge classification}
We first evaluated the performance of hypergraphlet kernels in the task of predicting the types of interactions between pairs of proteins in a PPI network, as well as interaction types and modes of action between proteins and chemicals in DTI data. As described in Section \ref{sec:problem} we first converted the input hypergraph to its dual hypergraph and then used the dual hypergraph for vertex classification. Table~\ref{edge:classification:results} lists the AUC estimates for each method and data set. Figure~\ref{auc:curves} shows ROC curves for one representative data set from each classification task and network type. Observe that the edit distance kernel ($\tau = 1$)  outperformed the traditional hypergraphlet kernel ($\tau = 0$) on all data sets. Edit distance kernels achieved the highest AUCs on two of the three data sets over random walk kernels. Therefore, these results provide evidence of the feasibility of this alternative approach to edge classification via exploiting hypergraph duality.

%\begin{table*}[t]
\begin{sidewaystable}[]
\begin{center}
\caption{\textbf{Area under the ROC curve estimates for each method on the PPI data sets using $10$-fold cross-validation.} The highest performance for each data set is shown in boldface.}
\label{ppi:lp:results}
\begin{tabular}{llllllllll}
\hline
\textbf{Method/Dataset} & \textbf{EC} & \textbf{SC} & \textbf{CE} & \textbf{AT} & \textbf{DM} & \textbf{HS} & \textbf{SP} & \textbf{RN} & \textbf{MM} \\
 \hline\hline
  \multicolumn{10}{c}{Without domain information, $|\Sigma| = 1$} \\ 
 Hypergraphlet kernel ($\tau = 0$) & $0.671$ & $0.878$ & \textbf{0.900} & $0.882$ & $0.846$ & $0.839$ & $0.809$ & $0.808$ & $0.843$ \\ 
 Hypergraphlet kernel ($\tau = 1$) & $0.659$ & $0.877$ & $0.895$ & $0.879$ & $0.844$ & $0.838$ & $0.804$ & $0.791$ & $0.834$ \\ 
  \multicolumn{10}{c}{With domain information, $\Sigma = \{\Sigma_{\text{SK}}, \Sigma_{\text{GO}}\}$ and $|\Sigma| = \{4, 8, 16\}$} \\ 
Random walk & $0.602$ & $0.501$ & $0.596$ & $0.626$ & $0.565$ & $0.531$ & $0.671$ & $0.647$ & $0.568$ \\
Cumulative random walk & $0.591$ & $0.527$ & $0.526$ & $0.632$ & $0.589$ & $0.620$ & $0.697$ & $0.656$ & $0.579$ \\
Pairwise spectrum kernel ($k = \{3,4,5\}$) & $0.612$ & $0.847$ & $0.827$ & \textbf{0.911} & $0.805$ & $0.877$ & $0.751$ & $0.771$ & $0.810$ \\
Hypergraphlet kernel ($\Sigma_{\text{SK}}; \tau = 0$) & $0.688$ & $0.869$ & $0.886$ & $0.880$ & $0.856$ & $0.843$ & $0.793$ & $0.797$ & $0.834$ \\
Hypergraphlet kernel ($\Sigma_{\text{GO}}; \tau = 0$) & \textbf{0.742} & $0.877$ & $0.887$ & $0.876$ & $0.854$ & $0.848$ & $0.802$ & $0.797$ & $0.831$ \\
Hypergraphlet kernel ($\Sigma_{\text{SK}}; \tau = 1$) & $0.707$ & $0.868$ & $0.885$ & $0.880$ & $0.857$ & $0.837$ & $0.797$ & $0.807$ & $0.835$ \\
Hypergraphlet kernel ($\Sigma_{\text{GO}}; \tau = 1$) & $0.740$ & $0.881$ & $0.887$ & $0.883$ & $0.854$ & $0.854$ & \textbf{0.830} & $0.801$ & $0.838$ \\
Hypergraphlet + Pairwise spectrum & $0.740$ & \textbf{0.883} & $0.893$ & $0.907$ & \textbf{0.858} & \textbf{0.878} & $0.825$ & \textbf{0.818} & \textbf{0.847} \\
\hline
\end{tabular}
\end{center}
%\end{table*}
\end{sidewaystable}

\begin{table}[t]
\begin{center}
\caption{\textbf{Area under the ROC curve estimates for each method on the DTI data sets using $10$-fold cross-validation.} The highest performance for each data set is shown in boldface.}
\label{dti:lp:results}
\begin{tabular}{llllll}
%\begin{tabular}{lp{0.56cm}p{0.56cm}p{0.56cm}p{0.56cm}}
\hline
\textbf{Method/Dataset} & \textbf{EZ} & \textbf{IC} & \textbf{GR} & \textbf{NR}  \\
 \hline\hline
 \multicolumn{5}{c}{Without domain information, $|\Sigma| = 1$} \\
Hypergraphlet kernel ($\tau = 0$) & $0.900$ & $0.837$ & $0.845$ & $0.895$ \\
Hypergraphlet kernel ($\tau = 1$) & $0.900$ & $0.838$ & $0.842$ & $0.889$ \\
 \multicolumn{5}{c}{With domain information, $\Sigma = \{\{\Sigma_{\text{SK}}, \Sigma_{\text{GO}}\} \bigcup \Sigma_{\text{SS}}\}$} \\
Random walk & $0.777$ & $0.704$ & $0.752$ & $0.835$ \\
Cumulative random walk & $0.919$ & $0.722$ & $0.765$ & $0.797$ \\
Hypergraphlet kernel ($\Sigma_{\text{SK}}; \tau = 0$) & $0.917$ & $0.838$ & $0.850$ & $0.918$ \\
Hypergraphlet kernel ($\Sigma_{\text{GO}}; \tau = 0$) & $0.916$ & $0.845$ & $0.850$ & $0.923$ \\
Hypergraphlet kernel ($\Sigma_{\text{SK}}; \tau = 1$) & $0.913$ & $0.845$ & $0.854$ & $0.933$ \\
Hypergraphlet kernel ($\Sigma_{\text{GO}}; \tau = 1$) & \textbf{0.922} & \textbf{0.863} & \textbf{0.858} & \textbf{0.941} \\
\hline
\end{tabular}
\end{center}
\end{table}

\subsection{Performance analysis on link prediction}

The performance of hypergraphlet kernels was further evaluated on the problem of link prediction on multiple PPI and DTI network data sets. Tables~\ref{ppi:lp:results} and~\ref{dti:lp:results} show the performance accuracies for each hypergraph-based method across all link prediction data sets. These results demonstrate good performance of our methods, with edit-distance kernels generally having the best performance. The primary objective of our study was to present a new approach whose value will increase as biological data becomes more frequently modeled by hypergraphs. At this time, such data sets are not readily available.

\subsection{Estimating interactome sizes}
We used the AlphaMax algorithm ~\cite{Jain2016b} for estimating class priors in positive-unlabeled learning to estimate the number of missing links and misannotated (false positives) interactions on each PPI network. For example, if we assume  a tissue and cellular component agnostic model (i.e., any two proteins can interact), we obtained that the number of missing interactions on the largest component of the human PPI network (see Table~\ref{datasets}) is about 5\% (i.e., approximately 2.5 million interactions), while the number of misannotated interactions is close to 11\% which translates to about 4,985 interactions. In the case of yeast, we computed that less than 1\% of the potential protein interactions are missing which is close to 95,000. The number of misannotated interactions is close to 13\%, which is about 3,400 misannotated protein pairs. Some of these numbers fall within previous studies that suggest that the size of the yeast interactome is between 13,500 ~\cite{Stumpf2008} and 137,000 ~\cite{Huang2007}; however, the size of the human interactome is estimated to be within 130,000 \cite{Kavitha2009} and 650,000 ~\cite{Stumpf2008} interactions. A recent paper by Lewis \emph{et al.}~\cite{Lewis2012} presents a scenario where yeast and human interactome size could reach 400,000 and over two million interactions, respectively. In any case, we note that these estimates were made as a proof of concept for the proposed methodology under the assumption of representative positive data. They however can serve as further validation of the usefulness of our problem formulation and underlying methodology. Additional tests and experiments, potentially involving exhaustive classifier and parameter optimization, will be necessary for more accurate and reliable estimates, especially for understanding the influence of potential biases within the PPI network data.

\section{Related work}

The literature on the similarity-based measures for learning on hypergraphs is relatively scarce. Most studies revolve around the use of random walks for clustering that were first used in the field of circuit design~\cite{Cong1991}. Historically, typical hypergraph-based learning approaches can be divided into (1) tensor-based approaches, which extend traditional matrix (spectral) methods on graphs to higher-order relations for hypergraph clustering~\cite{Cong1991, Rota2009, Leordeanu2012}, and (2) approximation-based approaches that convert hypergraphs into standard weighted graphs and then exploit conventional graph clustering and (semi-) supervised learning~\cite{Agarwal2005, Zhou2006}. The methods from the first category provide a direct and mathematically rigorous treatment of hypergraph learning, although most tensor problems are NP-hard. As a consequence, this line of research remains largely unexplored despite a renewed interest in tensor decomposition approaches~\cite{Hein2013, Purkait2014}. Regarding the second category, there are two commonly used transformations for graph-based hypergraph approximation: (1) the star expansion and (2) the clique expansion. These methods are reviewed and compared by Agarwal \emph{et al.}~\cite{Agarwal2006}. 

Under a supervised learning framework, Wachman and Khardon~\cite{Wachman2007} propose random walk-based hypergraph kernels on ordered hypergraphs, while Sun \emph{et al.}~\cite{Sun2008} present a hypergraph spectral learning formulation for multi-label classification. More recently, Bai \emph{et al.}~\cite{Bai2014a} introduced a hypergraph kernel that transforms a hypergraph into a directed line graph and computes a Weisfeiler-Lehman isomorphism test between directed graphs. A major drawback of most such approaches is that no graph representation fully captures the hypergraph structure. For instance, Ihler \emph{et al.}~\cite{Ihler1993} have shown that it is impossible to have an exact representation of a hypergraph via a graph while still retaining its cut properties. Therefore, there is a need for a robust hypergraph-based methodology for learning directly on hypergraph data.

\section{Conclusions}

This paper presents a learning framework for the problems of vertex classification, (hyper)edge classification, and link prediction in graphs and hypergraphs. The key to our approach is the use of hypergraph duality in order to cast each classification problem as an instance of vertex classification. This work also presents a new family of kernel functions defined directly on hypergraphs. Using the terminology of Bleakey \emph{et al.}~\cite{Bleakey2007}, our method belongs to the category of ``local'' techniques. That is, it captures the structure of local neighborhoods, rooted at the vertex of interest, and should be distinguished from ``global'' models such as Markov Random Fields or diffusion kernels~\cite{Kondor2002}. The body of literature on graph learning is vast. We therefore selected to perform extensive comparisons against a limited set of methods that are most relevant to ours.

The development of hypergraphlet kernels derives from the graph reconstruction conjecture, an idea of using small graphs to probe large graphs~\cite{Bondy1977,Borgs2006}. Hypergraphlet kernels prioritize accuracy over run time and, it may be argued, do not follow some recent trends in machine learning that generally trade off accuracy for improved scalability and real-time performance. We therefore propose that hypergraphlet kernel approaches, in particular those based on edit distances, be predominantly used on sparse graphs of moderate size. Fortunately, all graphs used in this work fall into that category. Increased accuracy, in general, benefits experimental biologists who typically use prediction to prioritize targets for experimental validation.

The proposed methodology was evaluated on multiple data sets for edge classification and link prediction in biological networks. The results show that hypergraphlet kernels are competitive with other approaches and readily deployable in practice. Through limited tests, we also find that combining hypergraphlet kernels with pairwise spectrum kernels achieves better accuracy than either of the methods does individually. 

\section{Acknowledgments}

We thank Matthew Carey for his help in implementing hyperedge-indel kernels. This work was partially supported by the National Science Foundation (NSF) grant DBI-1458477, National Institutes of Health (NIH) grant R01 MH105524, and the Indiana University Precision Health Initiative.

\bibliographystyle{plain}

\newpage
\section*{Appendix}
\subsection{Enumeration of labeled hypergraphlets}
\label{growth:pattern}

Here we characterize the feature space of fully labeled hypergraphlets by describing the dimensionality of count vectors  $\phi_{(n, \tau)} (v)$. We are interested in the order of growth of $\kappa(n, \Sigma, \Xi)$ as a function of $n$, $\Sigma$ and $\Xi$.

Suppose that $G$ and $H$ are base hypergraphlets with $n$ vertices and $m$ hyperedges. We say that $G$ and $H$ belong to the same equivalence class if and only if the total number of (non-isomorphic) fully labeled hypergraphlets corresponding to the base cases $G$ and $H$ are equal for any $\Sigma$ and $\Xi$. The total counts of labeled hypergraphlets over all alphabet sizes induce a partition of base hypergraphlets into equivalence classes. We denote the set of all equivalence classes over the hypergraphlets of order $n$ as $S(n)=\left\{S_{1}(n),S_{2}(n),\ldots\right\}$. For example, the set of vertex- and hyperedge-labeled $3$-hypergraphlets can be partitioned into either: two symmetry classes when $|\Xi| = 1$: $S_{1}(3) = \left\{3_2, 3_4, 3_7\right\}$ and $S_{2}(3) = \left\{3_1, 3_3, 3_5,3_6, 3_8, 3_9\right\}$, or seven symmetry classes when $|\Xi| > 1$: $S_{1}(3) = \left\{3_2\right\}$, $S_{2}(3) = \left\{3_4,3_7\right\}$, $S_{3}(3) = \left\{3_9\right\}$, $S_{4}(3) = \left\{3_6,3_8\right\}$, $S_{5}(3) = \left\{3_3\right\}$, $S_{6}(3) = \left\{3_5\right\}$ and $S_{7}(3) = \left\{3_1\right\}$. Table~\ref{hypergraphs:polya} summarizes equivalence classes induced by partitioning base hypergraphlets up to the order of $4$ along with the cardinality of each set. Overall, observe that the cardinality of $S(n)$ can be significantly larger than those reported for graphlets~\cite{Lugo2014} because the possible number of hyperedges in a hypergraphlet is generally much larger than the possible number of edges in a graphlet. Additionally, hyperedge-labels require base hypergraphlets $G$ and $H$ to have an equal number of hyperedges.

This approach can be generalized to hypergraphlets labeled by any alphabet $\Sigma$ and $\Xi$, such that
$$\kappa(n,\Sigma, \Xi)=\sum_{i=1}^{|S(n)|}m_{i}(n,\Sigma,\Xi)\cdot|S_{i}(n)|,$$

\noindent where $m_{i}(n,\Sigma,\Xi)$ is the number of (non-isomorphic) fully labeled hypergraphlets corresponding to any base hypergraphlet from the equivalence class $S_{i}(n)$. We use this decomposition to compute the total dimensionality of the count vectors by first finding the equivalence classes corresponding to the base hypergraphlets and then counting the number of labeled hypergraphlets for any one member of the group.

\begin{table}
\begin{center}
\caption{\textbf{Equivalence classes over vertex- and hyperedge-labeled hypergraphlets.} List of equivalence classes and their cardinality produced by partitioning the set of undirected base hypergraphlets for $n\in \{1,2,3,4\}$ over vertex-labels alphabet $\Sigma$ and hyperedge-labels alphabet $\Xi$. We also list the number of vertex-labeled $n$-hypergraphlets over an alphabet $\Sigma$ denoted as $m_{i}(n,\Sigma,1)$, as well as fully labeled $n$-hypergraphlets over an alphabet $\Sigma$ and $\Xi$ denoted as $m_{i}(n,\Sigma,\Xi)$.}
\label{hypergraphs:polya}
\begin{tabular}{llc}
\hline 
\multicolumn{3}{c}{\textbf{Vertex-labeled hypergraphlets}} \\\hline
$S_i (n)$ & $|S_i (n)|$ & $m_{i}(n,\Sigma,1)$ \\\hline\hline
$S_1(1)$ & $1$ & $|\Sigma|$ \\\hline
$S_1(2)$ & $1$ & $|\Sigma|^2$ \\
\hline
$S_1(3)$ & $3$ & $|\Sigma|^3$ \\
$S_2(3)$ & $6$ & $\frac{1}{2} (|\Sigma|^3 + |\Sigma|^2)$ \\
\hline
$S_1(4)$ & $221$ & $|\Sigma|^4$ \\
$S_2(4)$ & $212$ & $\frac{1}{2} (|\Sigma|^4 + |\Sigma|^3$) \\
$S_3(4)$ & $28$ & $\frac{1}{6} (|\Sigma|^4 + 3 \cdot |\Sigma|^3 + 2 \cdot |\Sigma|^2$) \\ \hline
\multicolumn{3}{c}{\textbf{Fully-labeled hypergraphlets}} \\\hline
$S_i (n)$ & $|S_i (n)|$ & $m_{i}(n,\Sigma,\Xi)$ \\\hline\hline
$S_1(1)$ & $1$ & $|\Sigma|$ \\\hline
$S_1(2)$ & $1$ & $|\Sigma|^2 \cdot |\Xi|$ \\
\hline
$S_1(3)$ & $1$ & $|\Sigma|^3 \cdot |\Xi|^{3}$ \\
$S_2(3)$ & $2$ & $|\Sigma|^3 \cdot |\Xi|^{2}$ \\
$S_3(3)$ & $1$ & $\frac{1}{2} (|\Sigma|^3 \cdot |\Xi|^{4} + |\Sigma|^2 \cdot |\Xi|^{3})$ \\
$S_4(3)$ & $2$ & $\frac{1}{2} (|\Sigma|^3 \cdot |\Xi|^{3} + |\Sigma|^2 \cdot |\Xi|^{2})$ \\
$S_5(3)$ & $1$ & $\frac{1}{2} (|\Sigma|^3 \cdot |\Xi|^{2} + |\Sigma|^2 \cdot |\Xi|^{2})$ \\
$S_6(3)$ & $1$ & $\frac{1}{2} (|\Sigma|^3 \cdot |\Xi|^{2} + |\Sigma|^2 \cdot |\Xi|)$ \\
$S_7(3)$ & $1$ & $\frac{1}{2} (|\Sigma|^3 \cdot |\Xi| + |\Sigma|^2 \cdot |\Xi|)$ \\
\hline
\end{tabular}
\end{center}
\end{table}

In the case of undirected fully labeled hypergraphlets, $m_{i}(n,\Sigma,\Xi)$ can also be computed by applying the theory of enumeration developed by P\'{o}lya~\cite{Polya1937}. In order to get the derivation of the complete generating function for each equivalence class $S_{i}(n)$, we first define the automorphism group $\mathcal{A}$ of a given vertex- and hyperedge-labeled hypergraph $G=(V,E)$. That is, in the case of fully-labeled hypergraphs, set $\mathcal{A}$ is a collection of permutations (automorphisms) of $V$ and $E$. Therefore, the counting problem can be re-formulated as follows: Let $G$ be a base hypergraphlet of $n$ vertices and $m$ hyperedges, and $\mathcal{A}$ be the automorphism group of $G$ over $V$ and $E$. Then, each permutation $\alpha \in \mathcal{A}$ can be written uniquely as the product of disjoint cycles such that for each integer $k \in \left\{1,\ldots,n\right\}$ ($k' \in \left\{1,\ldots,m\right\}$), we define $j_{k}(\alpha)$ ($j_{k'}(\alpha)$) as the number of cycles of length $k$ ($k'$) in the disjoint cycle expansion of $\alpha$. Interestingly, the generalized formula for the cycle index of $\mathcal{A}$, denoted as $Z(\mathcal{A})$, is a polynomial in $s_{1},\ldots,s_{n};s'_{1},\ldots,s'_{m}$ given by 
$$Z(\mathcal{A};s_{1},\ldots,s_{n};s'_{1},\ldots,s'_{m})=\frac{1}{|\mathcal{A}|}\sum_{\alpha\in \mathcal{A}}\prod_{k=1}^{n}\prod_{k'=1}^{m}s_{k}^{j_{k}(\alpha)} \cdot s_{k'}^{j_{k'}(\alpha)}.$$

\noindent By applying P\'{o}lya's theorem in the context of enumerating vertex- and hyperedge-labeled hypergraphlets corresponding to any base hypergraphlet in $S_i(n)$, we get that $m_{i}(n,\Sigma,\Xi)$ is determined by substituting $|\Sigma|$ for each variable $s_k$ and $|\Xi|$ for each variable ${s'}_{k'}$ in $Z(\mathcal{A})$. Hence,
$$m_{i}(n,\Sigma,\Xi)=Z(\mathcal{A};|\Sigma|, |\Sigma|, \ldots, |\Sigma|;|\Xi|, |\Xi|, \ldots, |\Xi|),$$

\noindent where $\mathcal{A}$ is the automorphism group of a base hypergraphlet from $S_{i}(n)$. As an example, consider the equivalence class $S_{3}(3)=\{3_9\}$ with $\Sigma = \{A, B, C\}$ and $\Xi = \{X, Y\}$ (Figure~\ref{hypergraphlets} illustrates an unlabeled version of hypergraphlet $3_9$). The automorphism group $\mathcal{A}=\{(v_1)(v_2)(v_3)(e_1)(e_2)(e_3)(e_4), (v_1)(v_2v_3)(e_1e_2)(e_3)(e_4)\}$; thus, $Z(\mathcal{A};s_{1},s_{2},s_{3};s'_{1},s'_{2}) = \frac{1}{2}(s_{1}^{3} \cdot {s'}_{1}^{4} + s_{1} \cdot s_{2} \cdot {s'}_{1}^{2} \cdot s'_{2})$. Therefore, it follows that, $m_{3}(3,\Sigma,\Xi) = Z(\mathcal{A};3,3,3;2,2) = \frac{1}{2}(3^{3} \cdot 2^{4} + 3 \cdot 3 \cdot 2^{2} \cdot 2) = 252.$

\end{document}